\documentclass[11pt,3p]{elsarticle}

\usepackage{lineno,hyperref}
\usepackage{verbatim} 

\usepackage{xcolor}
\usepackage{soul}

\begin{document}

\begin{frontmatter}

\title{On Deep Neural Networks for Detecting Heart Disease}


\author{Nathalie-Sofia Tomov\corref{cor1}}
\ead{ntomov@vols.utk.edu}
\author{Stanimire Tomov\corref{}}
\ead{tomov@icl.utk.edu}
\address{University of Tennessee, Knoxville, USA}


\cortext[cor1]{Corresponding author}

\begin{abstract}
    Heart disease is the leading cause of death, and experts estimate that approximately half of all heart attacks and strokes occur in people who have not been flagged as 'at risk.' Thus, there is an urgent need to improve the accuracy of heart disease diagnosis.
To this end, we investigate the potential of using data analysis, and in particular the design and use of deep neural networks (DNNs) for detecting heart disease based on routine clinical data. Our main contribution is the design, evaluation, and optimization of DNN architectures of increasing depth for heart disease diagnosis. This work led to the discovery of a novel five layer DNN architecture -- named {\it Heart Evaluation for Algorithmic Risk-reduction and Optimization Five} (HEARO-5) -- that yields best prediction accuracy. HEARO-5's design employs regularization optimization and automatically deals with missing data and/or data outliers. To evaluate and tune the architectures we use k-way cross-validation as well as Matthews correlation coefficient (MCC) to measure the quality of our classifications. The study is performed on the publicly available Cleveland dataset of medical information, and we are making our developments open source, to further facilitate openness and research on the use of DNNs in medicine. The HEARO-5 architecture, yielding 99\% accuracy and 0.98 MCC, significantly outperforms currently published research in the area.

\end{abstract}

\begin{keyword}
\texttt machine learning \sep DNN \sep cardiology \sep translational medicine \sep artificial intelligence \sep diagnosis \sep cardiovascular disease \sep diagnostic medicine \sep hyperparameter optimization
\end{keyword}

\end{frontmatter}


\section{Introduction}

Heart disease is the leading cause of death worldwide, killing twenty million people per year (WHO)\cite{who14}. An accurate and early diagnosis could be the difference between life and death for people with heart disease. However, doctors misdiagnose nearly 1/3 of patients as not having heart disease (British Medical Bulletin)~\cite{davies0101}, causing these patients to miss out on potentially life-saving treatment. This is a problem of increasing concern, as the number of Americans with heart failure is expected to increase by 46 percent by 2030 (American Heart Association)\cite{aha15}. 
What makes diagnosing heart disease a
challenging endeavor for any physician is that while chest pain and fatigue
are common symptoms of atherosclerosis, as many as 50 percent of people lack
any symptoms of heart disease until their first heart attack (Center for
Disease Control)~\cite{cdc16}. Discovering biomarkers (for heart disease) --
measurable indicators of the severity or presence of some disease --
is preferred~\cite{johann06}, but in many cases there are no clear biomarkers,
and multiple tests may be required and analyzed together.
Most doctors use the guidelines recommended by
the American Heart Association (AHA)~\cite{fisher17}, which test eight widely
recognized risk factors such as hypertension, cholesterol, smoking, and
diabetes~\cite{framingham,goff13}. However, this risk assessment model is flawed
since it rests on an assumed linear relationship between each risk
factor and heart disease outcome, while the relationships are complex
and with non-linear interactions~\cite{weng17}. Oversimplification may cause
doctors to make errors in their predictions, or overlook important factors
that could determine whether or not a patient receives treatment. Doctors
must also know how to interpret the diagnostic implications of medical
tests which vary between patients and require tremendous expertise.

The use of Machine Learning (ML) data analysis techniques
can alleviate the need for human expertise and the possibility of
human error while increasing prediction accuracy\cite{hutson17}. ML
algorithms apply flexible prediction models based on learned
relationships between variables in the input dataset. This can
prevent the oversimplification of fixed diagnosis models such as the
AHA guidelines. In fact, a neural network algorithm with 76 percent
accuracy has been proven to correctly predict 7.6\% more events than
the AHA method~\cite{paschalidis17}. Here, we significantly improve on these already
very promising ML results by designing and tuning deep neural network
(DNN) architectures of increasing depth for detecting heart disease
based on routine clinical data. We show that a flexible design and the
subsequent tuning of the (many) hyperparameters of a DNN can yield up to
99\% accuracy. The results were evaluated and validated using k-way
cross-validation as well as Matthews correlation coefficient (MCC) to
measure the quality of the classifications. The best results were obtained
on a novel five-layer DNN -- named {\it Heart Evaluation for Algorithmic
Risk-reduction and Optimization Five} (HEARO-5) -- that employs
regularization optimization and automatically deals with missing data
and/or data outliers. The HEARO-5's accuracy is 99\% with an MCC of 0.98,
which significantly outperforms currently published research in the area,
to further establish the appeal of using ML data analysis in diagnostic
medicine.

\section{Literature Review}
\label{sec:lit}

There are a number of research papers that use artificial neural networks to improve heart disease diagnosis. 
In a study published in the Journal of Cardiology, Yu et al. concluded that a neural network topology with two hidden layers was an accurate model with 94\% test data accuracy~\cite{yu}. They focus on the “multiplicity of risk factors” in constructing their model to classify features before determining a possible diagnosis. This study concluded that neural networks are an effective method of analyzing “cases when it is impossible to create a strict mathematical model but where there is a sufficiently representative set of samples.” Vinodhini et al. build on this research by performing feature classification with statistical models such as the chi square and then using a neural network as a predictive model~\cite{vinodhini}. This method proved successful overall, but exhibited weaker performance when given redundant attributes. A study published in the MHealth medical journal (Loh et al.) demonstrates the accuracy of deep neural networks by proving their ability to learn from nonlinear relationships in data~\cite{loh17}. However, they faced the issue of overfitting, when an algorithm learns too much from training data and becomes less capable of applying itself to unfamiliar data. Research published in the Journal of Healthcare Engineering helps address the problem of overfitting by ranking features, training the neural network with each feature ranking, and then training the neural network to output a potential diagnosis (Kim et al.)~\cite{kim17}. This helps ensure the network learns from numerically weighted important relationships in training data which can also be applied to unfamiliar data.

\section{Contribution to the Field}
\label{sec:contr}

While the current algorithms are effective, there is always a compelling need for improved algorithms that diagnose heart disease more accurately using accessible tests. 
To this end, the work described here makes the following main contributions:
\begin{itemize}
\item We designed of a unique and flexible heart disease diagnosis tool based on variable-layer DNN with regularization optimization that solely uses 
routine clinical data;
\item We developed HEARO-5 -- a specialized 5-layer DNN architecture for detecting heart disease, based on the evaluation and tuning of hyperparameters -- that is of very high accuracy (99\% and 0.98 MCC), significantly outperforming currently published research in 
the field;
\item The HEARO framework as a DNN data analytics research tool in diagnostic medicine and HEARO-5 as a benchmark, making them available for comparison and further studies, facilitating openness and research on the use of DNN techniques in medicine.
\end{itemize}

\section{DNN Background, Design, and Implementation}
\subsection{Overview}
The heart disease diagnostic tool that we designed uses standard 
fully-connected NNs. To construct the DNN architectures
for hearth disease diagnosis, one can use one out of the
many currently available frameworks, including
TensorFlow~\cite{abadi16}, Keras, PaddlePaddle,
Caffe, MagmaDNN~\cite{magmadnn}, etc. The most 
compute intensive building block for the DNN is the 
matrix-matrix multiplication (GEMM), which is available to use through highly optimized math libraries like CUBLAS, CUDNN, MKL, MAGMA~\cite{magmagemm},
and others. As the data that we have and need for training is not that large (see Section~\ref{dataset}), 
we developed a
parametrized DNN in Python, using NumPy as a backend for the linear algebra routines needed. The code is vectorized for performance, expressed in terms of matrix-matrix multiplications, and therefore can be easily ported to C/C++ code, calling highly optimized BLAS/GEMM implementations, which provides functional and performance portability across various computing architectures~\cite{magmadnn}. Our design and implementation is inspired by Andrew Ng's DNN designs and courses on deep learning, available on Coursera~\cite{coursera}.

\subsection{Flexible DNN design}
\begin{figure}[htbp]
  \centering
  \includegraphics[width=6in]{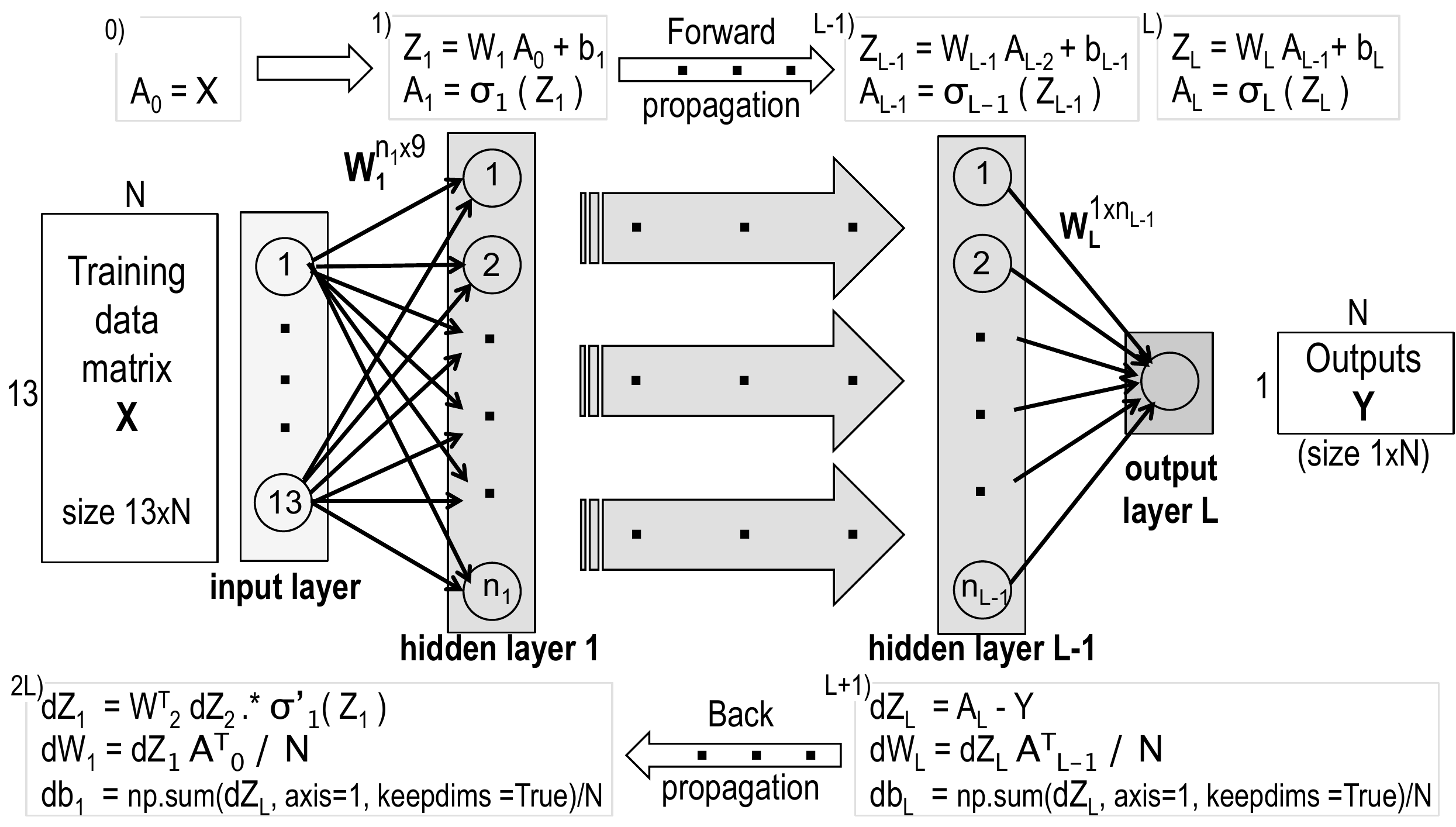}
  \caption{Parametrized DNN architecture and the main computational steps for its training. Training data $X$ consists of $13$ features (routine clinical data per patient; can also be parametrized) and $N$ training examples. Weights $W, b$ are trained using batch stochastic gradient descent method to make predictions $A_L$ "match" the given outcomes $Y$.}
  \label{fig:dnn-hearo}
\end{figure}

The framework design, notations, and main computational steps that we investigate are illustrated on Figure~\ref{fig:dnn-hearo}.
As shown, 
the neural network is organized into $L$ fully-connected 'layers' ($i = 1,..., L$) with $n_i$ nodes (or {\it artificial neurons}) per layer that function together to make a
prediction. The connections between layers $i-1$ and $i$
are represented by numerical weights, stored in matrix $W_i$ of size $n_i \times n_{i-1}$, and vector $b_i$ of length $n_i$. Thus, if the input values for layer $i$, given by the
values at the $n_{i-1}$ nodes of layer $i-1$, are represented as a vector $a_{i-1}$ of size $n_{i-1}$, the output of layer $i$ will be a vector of size $n_i$, given by the matrix-vector product $W_i a_{i-1} + b_i$. As training will be done in parallel for a
batch of {\it nb} vectors, the inputs $a_{i-1}$ will be matrices $A_{i-1}$ of size $n_{i-1} \times nb$ and the outputs will be given by the matrix-matrix products $Z_i = W_i A_{i-1} + b_i$, where "+" adds $b_i$ to each of the $nb$ columns of the resulting matrix.

\subsection{Main DNN building blocks}
The {\it Forward propagation} process, given by steps $0, ..., L$,  represents a non-linear hypothesis/prediction function $H_{W,b}(X) \equiv A_L$ for given inputs $X$ and fixed weights $W, b$. The weights must be modified so that the predictions $H_{W,b}(X)$ become close to given/known outcomes stored in $Y$. This is known as a {\it classification} problem and is a case of so called {\it supervised learning}. The modification of the weights is defined as a minimization problem on a convex cost function $J$, e.g.,
\[
        \min_{W, b} J (W, b), \textbf{ where }
        J(W, b) = -\frac{1}{N} \sum_{i=1}^{N} y_i \log H_{W,b}(x_i) + (1-y_i) \log (1 - H_{W,b}(x_i)). 
\]
This is solved by a batch stochastic gradient descent method -- an iterative algorithm using a batch of $nb$ training examples at a time. The derivatives of $J$ with respect to the weights ($W$ and $b$) are derived over the layers using the chain rule for differentiating compositions of functions. They are computed then by the {\it backward propagation} steps $L+1$, ..., $2L$, and used to modify their respective weights $W_i, b_i$ during the iterative training process for each layer $i$ as:
\[
  W_i = W_i - \lambda dW_i, ~~b_i = b_i - \lambda db_i,
\]
where $\lambda$ is a hyperparameter referred to as {\it learning rate}. The $\sigma_1, ..., \sigma_L$ functions are the activation functions (possibly different) for the different layers of the network, and $\sigma'$ are their derivatives. We have coded activation function choices for ReLU, sigmoid, tanh, and leaky ReLU. The ".*" notation is for point-wise multiplication.

\subsection{Algorithmic Optimization: Regularization}
    Regularization is a standard technique that prevents overfitting by penalizing large weight values.
DNNs tend to assign higher weight values for certain training data points, which corresponds to a high variance. Regularization helps address the problem of high variance on training data, which can improve accuracy on test data. Regularization is typically done by adding a penalty term of the form
$\frac{\alpha}{2 N} || W, b ||^2$
to the cost function $J$, where $|| W, b ||$ is some norm of the weights, e.g., L1 or L2.
The regularization parameter $\alpha$ imposes a penalty on large weights, thereby ensuring that we do not overfit training data. Another advantage of regularization is that it can prevent an algorithm from learning from data outliers, which is essential for a smaller dataset such as the heart disease patient set used in this research. 
Regularization causes the outliers to remain in the dataset, but reduces the algorithm's likelihood of learning from these values. Therefore, we add regularization to our model to investigate possible improvements in accuracy by reducing overfitting and automatically decreasing the impact of any outliers.

\subsection{Hyperparameter Optimization}
Part of the challenge of coding a neural network is structuring it so it is both accurate and efficient. One needs to determine how many layers to use, how many nodes per layer to use, etc. This was critical for example in deep convolutional networks for image recognition, where  significant improvement on the prior-art configurations was achieved by pushing the depth to 16-19 weight layers, e.g.,
as in the popular VGG network~\cite{vgg16}. Here also, we have determined that tuning for the depth and number of nodes per layer is critical for the accuracy. Moreover, we parametrized our framework, referred to as HEARO further on, as given above and in Figure~\ref{fig:dnn-hearo}. 

The parameters control the network configuration and accuracy, and therefore, the network must be highly optimized/tuned for them. These configuration parameters are also called {\it hyperparameters}. A HEARO configuration is determined by the following list of hyperparameters:
\begin{equation}\label{hparams}
     {\bf HEARO\_hparams} = [L,~ n1, ..., nL,
               ~\sigma 1, ..., \sigma L, 
               ~\lambda, ~\alpha, ~nb, ~epochs],
\end{equation}
where $epochs$ is the number of training iterations throughout the entire training set $X$, $\sigma i$ is the activation function for layer $i$ ($1$, $2$, $3$, or $4$ for ReLU, sigmoid, tanh, or leaky ReLU, respectively),  
and the rest are as given above. 

Thus, given a list of hyperparameters HEARO\_hparams, the HEARO framework trains itself (determining weights $W, b$) on a given input training data set $X$ and specified outcomes $Y$, and the challenge now becomes how to select the "best" hyperparameters.

\section{Optimization Methodology and the HEARO-5 Architecture}
\label{sec:desgn}

\subsection{Information About Dataset}\label{dataset}
HEARO uses training and test data from the University of California Irvine machine learning repository. Data have been preprocessed, where missing values are replaced with the value -1 to prevent them from significantly impacting the algorithm's model. There were approximately 12 missing values total. We applied also feature scaling to unit length.
This dataset, provided by the Cleveland Clinic Foundation, contains 75 total attributes of patient medical information for 303 patients\cite{dataset1}. The following 13 attributes are used: 1) age, 2) sex, 3) chest pain type, 4) resting blood pressure, 5) cholesterol, 6) fasting blood sugar, 7) resting electrocardiographic results, 8) maximum heart rate achieved, 9) exercise-induced angina, 10) ST depression, 11) slope of the peak exercise ST segment, 12) major vessels colored by fluoroscopy, and 13) thallium heart scan results. These attributes have been selected as optimal features by other researchers using this dataset~\cite{aravinthan} because they are considered most closely linked to heart disease.

Chest pain type is categorized by number, where 1 represents typical angina provoked by exercise or stress, and 2 represents atypical angina which is persistent chest discomfort~\cite{kawachi}. Common metrics such as resting blood pressure, cholesterol, and fasting blood sugar can be indicative of a patient’s general health and the state of their blood vessels, which is often shaped by the accumulation of plaque as an indicator of developing heart disease. Electrocardiogram results are visual representations of the heart’s activity, and can help doctors or algorithms determine if it is pumping at a normal rate or if circulation is impeded. An ST-T wave abnormality can be measured by wave height, and often has several implications: a ventricular aneurysm, coronary artery spasm, or artery tightness~\cite{davie}. These are all indicators of heart failure, and are therefore important features to an algorithm diagnosing heart disease. Slope of the peak exercise ST segment is a similar way of visually assessing the heart’s function when it must circulate more blood, in the case of exercise~\cite{lepeschkin}. In the dataset, this is characterized by the values 1, representing upward slope, 2, representing flat slope, and 3, representing downward slope. Thallium heart scans involve passing a radioisotope through the blood stream and visualizing where it reaches in the body. This can identify areas of the heart that are not receiving sufficient blood. Fluoroscopy is a similar evaluation tool that tracks the action of certain body parts, in this case the heart and nearby blood vessels. Vessels highlighted by fluoroscopy tests can indicate the presence of plaque accumulation. 

This dataset is used because it is publicly accessible and therefore improves the reproducibility of results. HEARO uses these thirteen features to diagnose heart disease because of their diversity, availability, and ability to identify heart disease at different stages of development. While some of the more detailed procedures such as fluoroscopy and thallium scans are often requested by a doctor for further information and are indicative of the presence/absence of heart disease, others such as blood sugar and cholesterol can provide tenuous evidence of abnormal activity~\cite{chitra17}. The combination of these features can create a model that accurately evaluates relationships between diverse patient conditions and heart disease diagnosis~\cite{ng17}.

\subsection{Accuracy Evaluation}\label{sec:accuracy}
In addition to measuring the percent accuracy, we also use K-fold cross validation to evaluate the accuracy. This is a standard technique for evaluating more accurately predictions, especially when the size of the training data set is small, like in our case. We use it also to flag possible cases of overfitting, which is often a threat when extensively tuning and the data set used is small. In order to maintain about a 2:1 training to test data ratio, we mostly use 3-fold cross validation, where two parts are assigned for training and one for testing.

Furthermore, we use the Matthews correlation coefficient (MCC) to analyze the algorithm's generalization abilities given a dataset with unbalanced class outcomes. 
In the Cleveland machine learning repository dataset, the class distribution of the two possible cases (0/1) is as follows: 164 '0' instances and 139 '1' instances. The MCC evaluates how well the algorithm performs on all possible data outcomes regardless of their ratio within the dataset. This adds further analysis to the potentially biased measure of percent accuracy. MCC is defined through the following formula: 
\[
       MCC = \frac{TP * TN - FP * FN}{\sqrt{(TP + FP)(TP + FN)(TN + FP)(TN + FN)}},
       \]
where TP represents true positives, TN represents true negatives, FP represents false positives, and FN represents false negatives.

\subsection{Optimization Methodology}\label{sec:tuning}
The HEARO framework gives us the flexibility to easily run and compare different configurations based on their accuracy tests described in Section~\ref{sec:accuracy}, which makes it a very good candidate for so called empirical optimization/tuning~\cite{li09}. This is a process where we generate a large number of possible configurations~(\ref{hparams}) and run them on a given platform to discover the one that gives the best results.

The effectiveness of empirical optimization depends on the chosen parameters to optimize, and the search heuristic used. A disadvantage is the time cost of searching for the best configuration variant, but in our case this is not a problem as the size of the data set is not that large. Furthermore, we restricted the search space by taking $L = 2..10$, $nL=1$, $ni=1..13$, $\lambda \in \{0.001, 0.01, 0.1\}$, L2 regularization with $\alpha \in \{0, 0.7, 1\}$, 
$nb = N$, and $epochs = 6000$.

Exhaustively testing the search space described above, using automated Python scripts to generate the different configurations and run them, revealed that the following configuration: 
\[
  {\bf HEARO\textbf{-}5} = [5, ~~9,7,5,3,1, ~~1,1,1,1,2, 
  	~~0.01, ~~0.7, ~~200, ~~6000 ]
\]
gives best accuracy, as further illustrated next in the results section.

\section{Results}
\begin{figure}[htbp]
  \centering
  \includegraphics[width=3.1in]{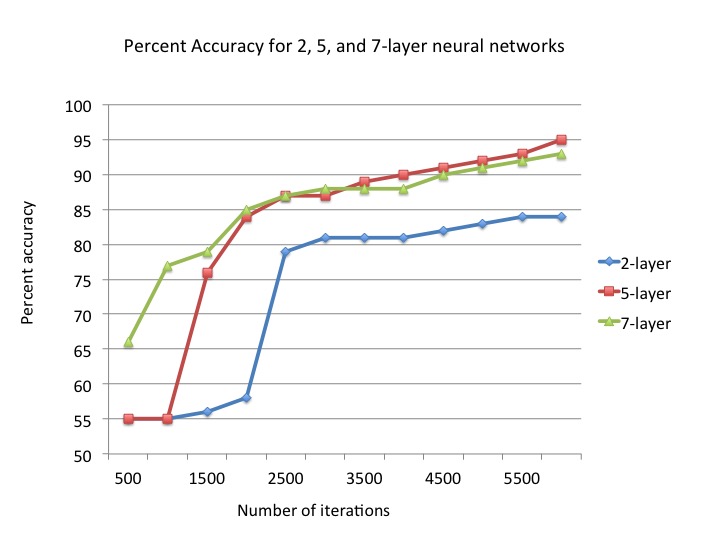}~~~
  \includegraphics[width=3.1in]{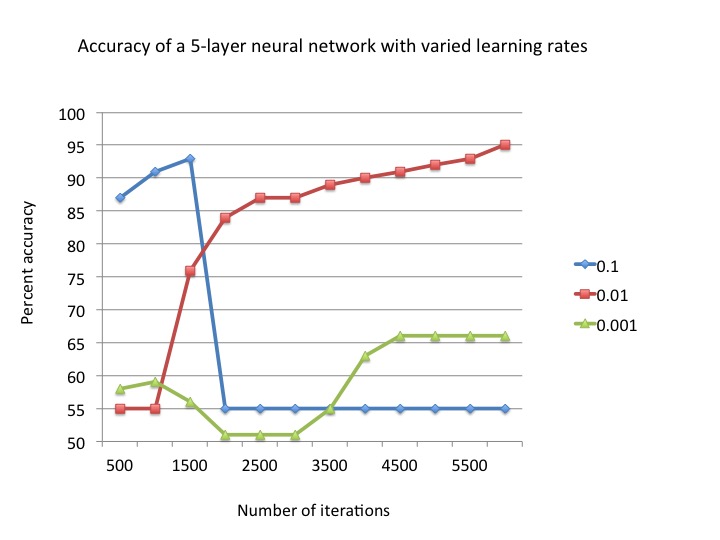}
  \caption{Left: Accuracy comparison of HEARO-5 {\it vs.} 
  	2 and 5 layer networks. Right: Effect of $\lambda$ 
    in HEARO-5.}
  \label{fig:testfig2}
\end{figure}

\begin{figure}[htbp]
  \centering
  \includegraphics[width=3in]{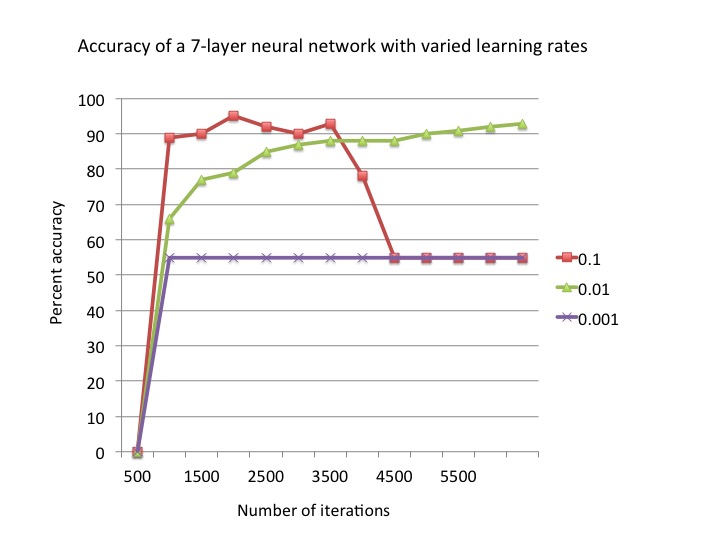}~~~
  \includegraphics[width=3in]{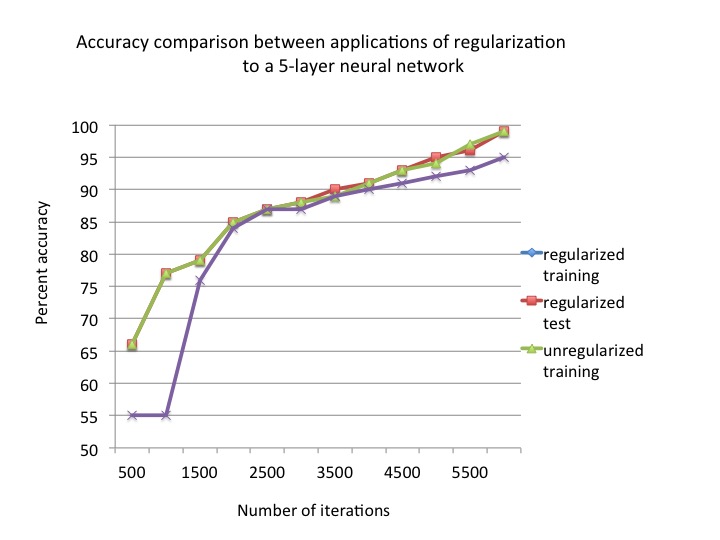}
  \caption{Left: Effect of $\lambda$ 
    on 7-layer HEARO architecture. Right: Effect on 
    regularization on HEARO-5 variants.}
  \label{fig:testfig3}
\end{figure}

The empirical optimization within the search space described in Section~\ref{sec:tuning} revealed that the HEARO-5 architecture yields best accuracy. 
The percent accuracy of HEARO-5 with $\alpha = 0$ (no regularization) is compared to the accuracy of configurations with 2 and 7 layers in Figure~\ref{fig:testfig2}, Left. All graphs indicate an algorithm's performance on test data unless specified otherwise. Figure~\ref{fig:testfig2}, Right
and Figure~\ref{fig:testfig3}, Left show the effect of the learning rate $\lambda$ on HEARO-5 and a 7-layer HEARO architecture, respectively.

HEARO-5 with $\alpha = 0.7$ exhibited 99\% accuracy on test data, and a Matthews correlation coefficient of 0.98. This is shown in Figure~\ref{fig:testfig3}, Right, and further discussed in Section~\ref{sec:regularization}.

\subsection{Comparison with Previously Published Results}
Stanford researchers used a convolutional neural network and obtained precision, recall and F1 scores of 0.80, 0.82, and 0.80 respectively~\cite{rajpurkar}. HEARO-5 outperforms these results, as it achieves precision, recall, and F1 of 0.98, 1, and 0.99 respectively.  
In 2016, Aravinthan et al. applied a Naive Bayes classifier and artificial neural network to this dataset with accuracy of 81.3\% and 82.5\%, respectively~\cite{aravinthan}. A study published in the International Journal of Computer Applications (Marikani) obtained results of 95.4\% and 96.3\% accuracy for classification tree and random forest algorithms~\cite{marikani}. 


Of interest is also to mention logistic regression
and the fact that it is the least accurate algorithm, 
likely due to its approach to fitting a fluctuating 
features data with non-linear correlation to heart 
disease.

    Furthermore on accuracy, our K-fold cross validation tests confirmed that HEARO-5 effectively reduces overfitting, as the cross validated accuracy was approximately the same as accuracy on test data with the set ratio. 
    
   The 0.98 MCC of HEARO-5 illustrates the HEARO-5 accurate evaluation of all class outcomes. Matthews correlation coefficient ranges from -1 to 1, where 1 represents perfectly balanced accuracy. Therefore, results of 0.98 MCC and 99\% accuracy are indicative of the algorithm's comprehensive data analysis model that is not skewed towards any particular outcome.

\subsection{Effect of Regularization}\label{sec:regularization}
While the unregularized DNN exhibits a discrepancy between training accuracy and test accuracy (99\% on training, 93\% on test), regularization increased the accuracy on test data to 99\%. 
The regularization improved the accuracy on test data by reducing the impact of outliers (and/or missing data) on training data.
On a relatively small dataset, outliers can inhibit the algorithm's ability to learn from consistent relationships in training data, and do not add scientific value. Therefore, by regulating the effect of outliers on learning, regularization improves the algorithm’s ability to generalize while maintaining the same scientific standard. Because regularization reduces overfitting on training data, the algorithm’s accuracy is expected to decrease on training data. In the case of HEARO-5 with $\alpha = 0$ (unregularized), a training accuracy of 99\% with somewhat lower test accuracy is indicative of overfitting (see Figure~\ref{fig:testfig3}, Right). The algorithm is learning from the ‘noise’ outlier values in the training data, detracting from its ability to generalize broader relationships in data. Learning these relationships is vital to accuracy on an unfamiliar dataset. 

\section{Conclusions and Future Directions}

This work investigated and showed the potential of using DNN-based data analysis for detecting heart disease based on routine clinical data. The results show that, enhanced with flexible designs and tuning, DNN data analysis techniques can yield very high accuracy (99\% accuracy and 0.98 MCC),
which significantly outperforms currently published research in the area, to further establish the appeal of using ML DNN data analysis in diagnostic medicine. 
Pending reviews and publication, we are preparing to release the HEARO software framework as an open source, and HEARO-5 as a benchmark, making the software available for comparison and further facilitating openness and research on the use of DNN techniques in medicine.


While the current developments are mostly research with excellent proof-of-concept results, further research and development is necessary in order to turn it into a robust diagnostic tool, e.g., that doctors consult and use routinely to make a more informed diagnosis.
Research is needed in the data analytics area and its intersection with data-based medical diagnosis -- including automatic search for best features, as well as possible features expansion or features reduction, e.g., due to lack of certain clinical data. Future directions include extending this analysis to construct a more thorough model that includes heart visualizations and CT image data.
More features can provide more data for the algorithm to learn from, creating a more complex model and ensuring a more accurate and detailed prediction. Another area of future research would involve using speed optimization tools and accelerated linear algebra backends such as 
MagmaDNN for GPUs to improve the algorithm's ability to process large amounts of data and find best configurations in parallel. In the future, HEARO
will also be developed into a production quality software package with a friendly user interface, e.g., to facilitate use by doctors or even patients directly.




\bibliographystyle{siamplain}

\end{document}